\documentclass[10pt]{article} 

\usepackage[accepted]{rlj} 

\usepackage{amssymb}            
\usepackage{mathtools}          
\usepackage{mathrsfs}           
\usepackage{graphicx}           
\usepackage{subcaption}         
\usepackage[space]{grffile}     
\usepackage{url}                
\usepackage{lipsum}             
\usepackage{wrapfig}

\usepackage{siunitx}
\DeclareSIUnit{\cent}{ct}

\title{The Challenges of Using Reinforcement Learning for Controlling Industrial Energy Systems}

\setrunningtitle{The Challenges of Using Reinforcement Learning for Controlling Industrial Energy Systems}

\author{Tobias Lademann\textsuperscript{1}, Théo Vincent\textsuperscript{2,3}, Jan Peters\textsuperscript{2,3,4}, Matthias Weigold\textsuperscript{1}}

\emails{t.lademann@ptw.tu-darmstadt.de}

\affiliations{
$^{1}$\textbf{Institute for Production Management, Technology and Machine Tools (PTW), Technical University of Darmstadt}\\
$^{2}$\textbf{DFKI GmbH, SAIROL}\\
$^{3}$\textbf{Department of Computer Science, Technical University of Darmstadt}\\
$^{4}$\textbf{Hessian.ai, Technical University of Darmstadt}\\
}


\summary{The summary.
\lipsum[1]
}

\begin{document}

\maketitle  

\vspace{-0.5cm}

\begin{abstract}
Reinforcement learning has shown promising results for optimizing the control of industrial energy systems, yet most existing studies remain limited to the application in simulation environments. We investigate the challenges of deploying reinforcement learning in a real-world industrial energy system, considering a thermal heating network as a use case. We formulate the task as a Markov Decision Process and systematically analyze the associated challenges along the structure of the formal description, including partial observability, action space design, reward design, and the simulation-to-reality gap. The challenges are grounded in an existing real-world deployment, where reinforcement learning achieves operational stability but shows a significant performance gap compared to simulation.
\end{abstract}

\vspace{-0.6cm}
\section{Introduction}

Industrial energy systems convert final energy such as electricity or natural gas into useful energy for production processes and building operation \citep{thiedeEnergyEfficiencyManufacturing2012}. Given their significant share of industrial energy demand \citep{rohdeErstellungAnwendungsbilanzenFur2021_2023}, improving the efficiency and flexibility of these systems while reducing operational costs is a key objective in industry \citep{thiedeEnergyEfficiencyManufacturing2012,thielDecarbonizeIndustryWe2021}.

Controlling industrial energy systems is a complex task due to interlinked thermal networks, multiple energy converters, the integration of energy storage systems, complex energy pricing models and strict requirements regarding security of supply \citep{kohneComparativeStudyAlgorithms2020,thiedeEnergyEfficiencyManufacturing2012}. Conventional control strategies are typically designed based on expert knowledge and implemented as static rule-based approaches, combining hysteresis controllers for discrete decisions with PID controllers for continuous control tasks \citep{frankFrameworkImplementationBuilding2024}. Although these approaches are generally robust, they are limited in their ability to define optimal control actions given a complex system behavior and conflicting optimization targets \citep{stavrevReinforcementLearningTechniques2024}.

To address these limitations, reinforcement learning (RL) has gained increasing attention in the energy system domain \citep{pereraApplicationsReinforcementLearning2021}, as it promises several advantages compared to conventional controllers, such as the ability to handle complex, uncertain and dynamic environments \citep{stavrevReinforcementLearningTechniques2024}. Existing studies demonstrate the potential of RL to improve the control of industrial energy systems. However, the application of RL in real-world environments remains limited: Most existing studies evaluate RL-based control strategies exclusively in simulation \citep{ranzauApplicationDeepReinforcement2025a}, neglecting the real-world challenges in an industrial setting \citep{dulac-arnoldChallengesRealworldReinforcement2021}. 

This paper addresses this gap by investigating the challenges associated with the real-world deployment of RL for controlling an industrial energy system. It can be seen as an open call for RL methods to tackle application-relevant challenges, thereby guiding RL researchers in framing their research problems.

\textbf{Contributions.}
(1) We model the control task of a representative industrial use case as an RL problem, tailored to the characteristics and constraints of real-world industrial applications. (2) We formulate challenges within the RL problem formulation. (3) We concretize the identified challenges based on a real-world deployment of an RL control strategy in the considered use case.

\section{Use case}
\label{sec:Use case}

The \textit{ETA Research Factory} serves as a real-world testbed for the deployment and evaluation of optimized control strategies under conditions representative of industrial energy systems. The energy system of the factory consists of three thermal networks, operating at different temperature levels \citep{frankFrameworkImplementationBuilding2024}. For this work, we consider the \textit{Heating Network High Temperature}, as it captures key characteristics of industrial energy systems, including multiple energy converters, thermal storage, and coupled energy carriers \citep{frankFrameworkImplementationBuilding2024}. At the same time, it is the only thermal network that enables reproducible real-world experiments, as ambient influences are negligible \citep{lademannRealWorldBenchmarkingControl2026}. The use case offers realistic industrial complexity and safety constraints while reducing experimental costs and risks compared to productive industrial environments.

\begin{figure}[ht]
    \begin{center}
        \includegraphics[width=0.9\textwidth]{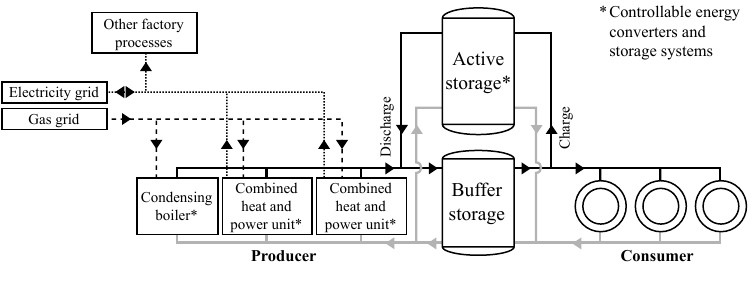}
    \end{center}
    \vspace{-0.8cm}
    \caption{Simplified scheme of the use case system based on \cite{frankHydraulicSchemeETA2025}.}
    \label{fig:hydraulic_scheme}
\end{figure}

The considered system consists of multiple energy converters and thermal storage components that jointly supply thermal power to production processes and building operation as shown in \autoref{fig:hydraulic_scheme}. In particular, it includes \textbf{two combined heat and power units} and \textbf{one condensing boiler}, which convert gas into thermal energy. The combined heat and power units additionally generate electricity, which is either used within the factory or fed into the grid, depending on the factory demand. Both reduce operating costs, depending on current electricity market prices and feed-in tariffs.

Thermal energy can be stored in an \textbf{active storage} system, allowing the temporal decoupling of energy production and consumption. Depending on the operating mode, the storage acts either as a thermal producer (discharge) or consumer (charge). This flexibility enables the operation of the combined heat and power units in periods of favorable electricity prices. The active storage is equipped with vacuum insulation and a layering system to reduce thermal and exergy losses. In addition, a passive \textbf{buffer storage}, which cannot be directly controlled, hydraulically decouples systems for thermal energy production from thermal energy consumption. 

On the demand side, the system supplies both production processes and building processes within the factory. All relevant energy flows, including thermal, electrical, and gas power, are measured for each energy converter and consumer. In addition, each storage is equipped with three temperature sensors located at the lower, middle, and upper parts of the storage.

\vspace{-0.2cm}

\vspace{-0.2cm}

The main objective of the system is to reliably supply thermal power to consumer systems at the required temperature level. The target temperature depends on the overall operating mode, with \qty{70}{\degreeCelsius} in production mode and \qty{55}{\degreeCelsius} in building mode. To ensure safe operation, the middle buffer storage temperature must remain within predefined limits, ranging from \qty{45}{\degreeCelsius} to \qty{85}{\degreeCelsius} in production mode and from \qty{40}{\degreeCelsius} to \qty{85}{\degreeCelsius} in building mode. 

RL has been applied to industrial energy systems in simulation and real-world settings, but deployment challenges remain largely unaddressed, as discussed in \autoref{appendix:Related Work}.

\section{Problem formulation}

\vspace{-0.4cm}

This section provides a general formulation of the control of the considered industrial energy system as an RL problem. For this purpose, the control task is described with a Markov Decision Process (MDP) \citep{suttonReinforcementLearningIntroduction2018}, where an agent interacts with the real-world environment. Given a state $s \in  \mathcal{S}$ and an action $a \in \mathcal{A}$, the environment transitions to a successor state $s' \in \mathcal{S}$ according to the transition probability $p(s' \mid s, a)$. A deterministic reward function of the state, action and next state is received after each transition. The agent learns a policy $\pi(a \mid s)$ that maps system states to action distributions. 

\textbf{State Space:} A reasonable state space should include all observable system states and exogenous inputs that influence the system behavior and optimal control decisions. It should also include forecasts of relevant exogenous inputs, such as thermal demand of consumer systems, electrical demand of other factory processes, and energy prices, to enable optimal control decisions. 

The state space is constructed from available sensor measurements acquired and provided by the building automation system (e.g. storage temperatures or heat flow rates) and exogenous inputs provided through external data interfaces (e.g. market prices). The state space can include both continuous (e.g. temperature) and discrete (e.g. energy converter state on/off) values.

\textbf{Action Space:} The RL agent controls the energy converters and active storage (see \autoref{fig:hydraulic_scheme}). Each component is controlled by a boolean activation variable and a continuous setpoint, whose interpretation and value range depend on the component \citep{fuhrlander-volkerModularDataModel2022}. The setpoint only affects the component when it is active. 

The RL agent operates at a supervisory control level and interacts with the real-world system through a control interface on the programmable logic controller (PLC). The RL actions overwrite the default rule-based control implemented on the PLC. A fallback mechanism ensures safe operation by restoring the rule-based controller if the buffer storage temperature limits defined in \autoref{sec:Use case} are violated. As only the supervisory control is overwritten, the local control implemented on the PLC remains active and continues to handle actuator sequencing (e.g. pumps and valves) and low-level control tasks such as temperature control \citep{fuhrlander-volkerModularDataModel2022}.

\textbf{Transition Dynamics:} The system state transitions according to the thermal, hydraulic and electrical dynamics of the system. State transitions of the physical system are influenced by the control actions, thermal demand, and the local control logic implemented in the building automation system. In addition, exogenous variables such as energy prices and production modes evolve independently of the agent actions according to external influences.

\textbf{Reward:} The performance of a control strategy can be evaluated using several evaluation metrics addressing different objectives. These objectives are partially conflicting, requiring the control strategy to balance trade-offs between economic, technical, and environmental performance \citep{poschGanzheitlichesEnergiemanagementFuer2011}. The following evaluation metrics are relevant for the presented use case, as defined by \cite{lademannRealWorldBenchmarkingControl2026}:

\begin{enumerate}[label=\Alph*.]
    \item \textbf{Operating costs} comprise electricity costs, gas costs, and maintenance costs. Electricity-related costs include expenses for external electricity purchase under a tariff based on the day-ahead market, as well as revenues from self-consumption and feed-in of locally generated electricity by the combined heat and power units.
    \item \textbf{Security of supply} is essential to guarantee that thermal demands of production and building systems are reliably met at all times. The security of supply is evaluated based on the mode-dependent temperature target \qty{55}{\degreeCelsius} or \qty{70}{\degreeCelsius} and the actual upper buffer storage temperature. Only deviations below the target temperature are problematic.
    \item \textbf{Energy efficiency} is considered to minimize overall energy consumption and improve the effective use of available energy resources. It is calculated as the ratio of total useful energy output, including thermal and electrical energy, to the total energy input from gas and electricity.
    \item \textbf{System wear} is considered to prevent excessive component degradation caused by frequent switching. For the combined heat and power units, the manufacturer specifies a minimum mean runtime of \qty{3}{\hour}. In general, longer runtimes are beneficial, as they reduce start-stop cycles, wear, and associated early component degradation. Additionally, the combined heat and power units are subject to a maximum allowable return temperature of \qty{65}{\degreeCelsius}.
    \item \textbf{CO$_2$ emissions} are considered to reduce the environmental impact of system operation and support decarbonization goals. 
\end{enumerate}

\textbf{Terminal States:} There are no terminal states. The building automation system enforces admissible temperature ranges and temporarily overrides the RL control actions with rule-based fallback control if these limits are violated. However, this does not terminate the deployment, as the RL control resumes once the system state returns to the admissible operating range.

\vspace{-0.4cm}

\section{Challenges}
\label{sec:Challenges}

\vspace{-0.5cm}

This section discusses challenges related to the problem formulation that arise when deploying RL in the considered real-world system. For clarity, the challenges are presented following the structure of the previously introduced MDP formulation.

\textbf{System State:} The system state is only partially observable, as it is often the case for real-world systems \citep{dulac-arnoldChallengesRealworldReinforcement2021,xiangRecentAdvancesDeep2021}. In particular, the state of charge of the storage systems, especially the active storage, cannot be accurately determined due to strong temperature stratification and the limited number of temperature sensors. While the presented use case provides extensive sensor information for all components, this level of observability is typically not available in industrial systems \citep{thiedeEnergyEfficiencyManufacturing2012}.

Furthermore, relevant future state information, such as forecasts of thermal or electrical demand depend heavily on production-related-factors and may be unavailable or uncertain \citep{waltherHierarchicalElectricalLoad2022}. Additional uncertainty arises from sensor inaccuracies or malfunctions. 


\textbf{Action Space:} The design of the action space is critical for real-world deployment, as industrial energy systems offer multiple possible ways to formulate control actions. While the action space must be compatible with the existing building automation interface, its formulation can still include several design choices, for example removing redundant actions, combining multiple control signals into one action, or discretizing continuous control variables \citep{kanervistoActionSpaceShaping2020}.

A priori, it is not clear which formulation facilitates learning most effectively and leads to the best performance. Integrating domain knowledge into the action space formulation can speed up the learning process. However, it also restricts the agent's degrees of freedom and may exclude the true optimal solution \citep{kanervistoActionSpaceShaping2020}, thereby limiting the achievable performance gains compared to a rule-based controller that already incorporates substantial inductive bias.

\textbf{Transition Dynamics:} All works discussed in \autoref{appendix:Related Work} rely on a simulation model to obtain transition dynamics during training. However, the implementation and validation effort for high-fidelity simulation models is substantial. Moreover, detailed information on the local controller logic and component behavior is often not available from manufacturers. 

This issue is further intensified by the fact that industrial energy systems are subject to continuous changes, ranging from gradual efficiency degradation of energy converters to major system modifications such as additional components or changed energy pricing models. Such changes may require repeated adaptation of the simulation model and retraining of the RL agent, which generally includes new hyperparameter searches.

Furthermore, detailed simulation models can result in stiff differential-algebraic equations and long simulation times due to the coupling of dynamics with different time constants, such as slow thermal dynamics and discrete switching operations \citep{blumBuildingOptimizationTesting2021}. As a result, RL training and especially hyperparameter optimization become computationally expensive.

\textbf{Reward:} In industrial energy systems, relevant objectives are often conflicting (e.g. operational costs and security of supply) and must be translated into a reward function that reflects the priorities of the system operator. 

A common approach is to formulate the reward as a weighted sum. As proper weighting of multiple objectives is essential \citep{schaferCrucialRoleProblem2025}, the reward design should reflect realistic operational priorities from the perspective of a system operator. However, many operational objectives, such as security of supply, energy efficiency, or system wear, cannot be directly quantified in monetary terms and therefore require relative weighting within the reward design. 

Additionally, rewards are often delayed, particularly for operational costs when storage systems are used to shift energy production and consumption over time to exploit periods of favorable electricity prices.

\textbf{Agent's Objective:} The objective of the RL agent can be formulated in different ways, including finite-horizon, average-reward, or discounted-return objectives \citep{suttonReinforcementLearningIntroduction2018}. A finite-horizon setting is not suited as the system is operated continuously. An average-reward setting appears favorable, as it mitigates the short-sighted effect of the discounted-return setting. However, this setting is still largely unexplored. Finally, the discounted setting requires specifying a discount factor, which should be tuned to balance the trade-off between capturing delayed rewards and prioritizing immediate rewards due to future uncertainty.

Since operational costs are strongly influenced by day-ahead market prices, the agent's objective should capture at least daily electricity price cycles. In addition, weekly factory demand patterns should be considered, as thermal and electrical demand typically differ between weekdays and weekends. However, increasing uncertainty in future states, thermal and electrical demand forecasts, and market conditions favors a stronger emphasis on immediate rewards.

Practical RL training and evaluation are usually performed in an episodic setting, requiring the definition of terminal states, suitable time horizons, and realistic and diverse initial state distributions \citep{schaferCrucialRoleProblem2025}. These design choices can strongly affect learning and performance \citep{schaferCrucialRoleProblem2025}.

\textbf{Control Frequency:} The control frequency must be high enough to capture the relevant system dynamics. However, thermal systems typically exhibit slow dynamics compared to other control tasks such as power grids \citep{schlegel2025towards}. High control frequencies lead to longer planning horizons, which can make the control problem harder.

Furthermore, if a simulator is used, the control frequency can be constrained by the computational effort of the simulation model, as smaller communication step sizes may force a variable-step solver to decrease its step size, which significantly increases runtime.

\textbf{Explainability:} Industrial deployment requires not only good performance but also system operator acceptance. Explainable RL can support the interpretation of learned policies \citep{bekkemoenExplainableReinforcementLearning2024}. However, the challenge is broader than explaining individual control actions. Even a well-performing RL agent may show operating patterns that differ substantially from a well-known rule-based baseline, which can reduce operator acceptance. 

\vspace{-0.3cm}

\section{A Study of Real-World Application}
\label{sec:Experiments}

\vspace{-0.2cm}

This section analyzes the real-world deployment of RL in the considered system by \cite{lademannRealWorldBenchmarkingControl2026} and links the formulation choices to the challenges we identified. Again, we present the application following the MDP structure.

\textbf{State Space:} The selected state space comprises all available storage temperatures (three per storage), the aggregated thermal demand of the consumer systems, the current operational state of the controlled systems (on/off;charge/discharge), the current runtime of the combined heat and power units, the production and heating mode of the factory, the electrical demand of other factory processes as well as time-dependent energy prices and remunerations. Although these signals would be available in the considered system, the authors do not include forecasts, thermal power of producer systems, or electrical and gas energy consumption in the state space, reflecting the partial observability typically encountered in industrial systems with limited sensor availability.

\textbf{Action Space:} The authors address the challenge of redundant actions by deriving a discrete action space from the rule-based control strategy. For each energy converter $i$, activation and temperature setpoints are combined into an action $a_i \in \mathcal{A}_i = \{0,1,2\}$, denoting off and operation at two predefined setpoints. For the active storage, the same action set represents discharge, idle, and charge at a fixed minimum setpoint. The overall action space is $\mathcal{A} = \mathcal{A}_1 \times \dots \times \mathcal{A}_4$. Thereby, the continuous setpoints are discretized to improve operational stability in real-world deployment, but this also limits the agent's degrees of freedom and, therefore, its optimization potential.

In addition, the authors use an action mask for the combined heat and power units to enforce a minimum runtime of \qty{3}{\hour}, further supporting operational stability in deployment.

\textbf{Transition Dynamics:} The system is modeled in \textit{Modelica} \citep{mattssonModelicaInternationalEffort1997} using the \textit{Thermal Systems Control Library} \citep{borstThermalSystemsControlLibraryModelicaLibrary2023}, a library dedicated for developing optimized control strategies. The model includes physical component models as well as supervisory and local control logic, thereby representing the real-world system behavior and control interface.

Additionally, the simulation model is calibrated using real-world data to reduce the simulation-to-reality (sim-to-real) gap. The authors report an absolute relative error between simulated and measured accumulated energy of \qty{6}{\percent} for electrical energy and \qty{5}{\percent} for gas energy over a two-day validation horizon. The calibration results further indicate unresolved model mismatch or measurement errors, as several calibrated parameters converge to the boundaries of their predefined ranges. This highlights the modeling and validation effort required when RL training relies on simulation-based transition dynamics.

\textbf{Reward:} The authors address the reward-design challenge by formulating the reward as a weighted sum of monetary and non-monetary cost terms. Monetary operating costs include electricity, gas, and maintenance costs, while non-monetary reward terms include temperature deviations, switch-on operations, minimum runtime violations, and episode termination.

The resulting weighting challenge is addressed by weighting only the non-monetary cost terms relative to the monetary terms. The rule-based strategy is used as a reference to obtain realistic and justifiable weights.

\textbf{Agent's Objective:} The authors chose a discounted return. To cope with the challenge of defining a discount factor, the authors include it in the hyperparameter search space and chose the undiscounted return as an evaluation metric. 

The training horizon is set to \qty{7}{\day} to capture daily and weekly variations in electricity prices as well as thermal and electrical demand. The evaluation horizon depends on the experiment setup. 

To obtain a diverse initial state distribution, the authors use a pseudo-random generator to sample from the available training data, while storage temperatures are initialized from predefined temperature ranges. In real-world evaluation, the initial state cannot be controlled and is determined by the actual operating conditions at the experiment start.

\textbf{Control Frequency:} The control frequency  is set to \qty{180}{\second}, favoring fast simulation times and a simplified control problem. However, whether this frequency is sufficiently high to fully capture the relevant system dynamics is not further analyzed.

The authors implement the experiments within the open-source Python framework \textit{ETA Factory Thermal System Operation} \citep{lademannETAFactoryThermal2026} using proximal policy optimization with invalid action masking \citep{huangCloserLookInvalid2022}. \autoref{tab:results_rule_rl} compares the rule-based and RL controllers in simulation and real-world deployment. Overall, the RL controller demonstrates operational stability in both simulation and real-world deployment.

Both controllers are evaluated under identical scenarios and initial states in simulation. In the real-world deployment, the scenario conditions (production/building mode, thermal demand of consumer systems and electrical power of other factory processes) are controlled as well. However, the initial states differ between real-world experiments, highlighting the challenge of comparability in real-world deployment using the undiscounted return.

To address this, the authors propose metrics A--E, whose normalization enables comparison despite differing initial states and time horizons. Performance is evaluated without termination to reflect realistic operation, since RL control resumes once the temperature returns to the admissible range after a fallback intervention. Temperature violations are therefore captured by the security of supply metric.

\vspace{-0.2cm}

\begin{table}[htbp]
    \caption{Performance evaluation of rule-based and RL control strategy over horizon $T$. The arrows indicate whether the value should be maximized ($\uparrow$) or minimized ($\downarrow$).}
    \vspace{-0.5cm}
    \begin{center}
        \begin{tabular}{llcccc}
            \hline
            & & \multicolumn{2}{c}{\bf Simulation ($T = \qty{90}{\day}$)} & \multicolumn{2}{c}{\bf Real-world ($T = \qty{3}{\day}$)} \\
            \cline{3-6}
            \bf Metric & 
            \bf Unit & 
            \bf Rule-Based & 
            \bf RL &
            \bf Rule-Based & 
            \bf RL \\
            \hline
            A. Operating costs ($\downarrow$) & \unit{\cent/\kilo\watt\hour} & 8.8 & \textbf{7.2} & \textbf{8.3} & 8.6 \\ 
            B. Security of supply ($\downarrow$) & \unit{\kelvin} & 1.1 & \textbf{0.9} & \textbf{1.1} & 3.5 \\ 
            C. Energy efficiency ($\uparrow$) & \unit{\percent} & 69.7 & \textbf{72.1} & \textbf{74.7} & 72.4 \\ 
            D. System wear ($\uparrow$) & \unit{\hour} & \textbf{15.6} & 4.4 & \textbf{11.2} & 10.8 \\ 
            E. CO$_2$ emissions ($\downarrow$) & \unit{\text{\gram\si{CO_2}}/\kilo\watt\hour} & 266.2 & \textbf{242.3} & 279.6 & \textbf{275.8} \\ 
            Sum of reward ($\uparrow$) & -- & -1839 & \textbf{-1345} & \textbf{-95} & -297 \\
            \hline
        \end{tabular}
    \end{center}
    \label{tab:results_rule_rl}
\end{table}

\vspace{-0.5cm}

In both simulation and real-world deployment, the agent permanently activates the active storage and frequently switches between charging and discharging, while the second combined heat and power unit remains completely inactive. This behavior supports operational stability by increasing the effective thermal inertia of the system, but does not exploit volatile energy market prices. We attribute this primarily to three coupled issues: first, the absence of forecasts in the state space limits the agent's ability to anticipate favorable market conditions; second, the reward weighting and delayed returns may not sufficiently support long-term cost optimization; and third, the discount factor underweights future rewards. Moreover, this behavior differs fundamentally from the rule-based strategy, which may reduce operator acceptance. 

In real-world deployment, the performance of RL is less promising than in simulation with respect to the metrics A--D. The RL controller shows a significantly larger deviation in security of supply, resulting in a substantially lower sum of rewards compared to the rule-based controller. Although a calibrated simulation model based on a dedicated simulation library is used, the RL performance achieved in simulation does not fully transfer to the real-world system. This highlights the challenge of simulation-based transition dynamics.

To summarize, the studied real-world deployment demonstrates how practical design choices can mitigate several of the identified challenges. However, the authors do not address all challenges, highlighting opportunities for further development and applications.

\section{Limitations, Conclusion \& Future Work}
\label{sec:Limitations}

This study only discusses the challenges of real-world RL deployment based on a single industrial energy system. Therefore, effects caused by other system topologies, energy converters, storage technologies, or demand structures are not covered. Studying different use cases could reveal more challenges.

A further limitation is the selected use case itself. The ETA Research Factory is a research testbed that enables controlled real-world deployment and representative safety constraints. However, additional technical, organizational, and economic challenges are expected when transferring RL-based control to productive industrial systems.

To conclude, we provide a structured discussion of the challenges associated with formulating and deploying RL for real-world industrial energy systems. The analysis shows that many challenges arise at the level of the RL problem formulation, including partial observability, action space design, reward weighting, delayed rewards, and the dependence on simulation-based transition dynamics. Grounded in an existing real-world deployment, we demonstrate how the identified formulation challenges arise in practice. Thus, our contribution can be seen as a starting point for developing RL to address application-relevant challenges.

In future work, we will investigate offline RL to reduce the dependence on high-fidelity simulation models and make better use of available real-world operational data \citep{levineOfflineReinforcementLearning2020}. In addition, online learning in the considered use case appears feasible, but requires methods for safe exploration and high sample efficiency \citep{guReviewSafeReinforcement2024,dulac-arnoldChallengesRealworldReinforcement2021}.


\subsubsection*{Acknowledgments}

The authors thankfully acknowledge the financial support of the project “ENIPRO“ (grant no. 03EN4111A) by the Federal Ministry for Economic Affairs and Energy (BMWE) and project supervision by the project management organization Projektträger Jülich (PtJ). This research was supported by “Third Wave of AI”, funded by the Excellence Program of the Hessian Ministry of Higher Education, Science, Research and Art, and by the grant “Einrichtung eines Labors des Deutschen Forschungszentrum für Künstliche Intelligenz (DFKI) an der Technischen Universität Darmstadt”. We gratefully acknowledge support from the hessian.AI Service Center (funded by the Federal Ministry of Education and Research, BMBF, grant no. 01IS22091) and the hessian.AI Innovation Lab (funded by the Hessian Ministry for Digital Strategy and Innovation, grant no. S-DIW04/0013/003).




\bibliography{main}
\bibliographystyle{rlj}

\appendix

\section{Related Work}
\label{appendix:Related Work}

RL has been applied to the control of industrial energy systems in both simulation and real-world applications comparable to the considered use case. However, existing research primarily focuses on the implementation and benchmarking of RL, while challenges related to real-world deployment remain largely unaddressed.

RL has been applied to various industrial energy systems including cooling supply systems for buildings and process cooling \citep{schreiberApplicationTwoPromising2020}, multi-energy industrial parks \citep{zhuEnergyManagementBased2022}, cooling systems of an industrial production site \citep{weigoldMethodApplicationDeep2021}, or cooling supply systems in the chemical and pharmaceutical industry \citep{lademannDeepReinforcementLearning2026}. RL has also been applied in the building energy sector \citep{wangReinforcementLearningApproach2023,hePredictiveControlOptimization2023}, which exhibit comparable network topology, although building energy systems are typically less complex regarding control complexity and safety requirements. While the presented studies demonstrate the potential of RL, all implementations are exclusively in simulation environments, with real-world deployment challenges remaining an open problem.

A very limited number of studies address real-world deployment of RL in industrial energy systems \citep{ranzauApplicationDeepReinforcement2025a}. \cite{ranzauApplicationDeepReinforcement2025a} deploys RL in the real-world system of the ETA Research Factory and develops methods for improving the sim-to-real transfer. \cite{lademannRealWorldBenchmarkingControl2026} consider the use case described in \autoref{sec:Use case} to develop common evaluation metrics for different control strategies and benchmark the performance of RL and other controllers in real-world application. Real-world RL applications have also been demonstrated in related domains, such as building climate control \citep{duDemonstrationIntelligentHVAC2022} and industrial battery control \citep{yiIntegratedEnergyManagement2022}. Again, these application areas exhibit less complexity compared to industrial energy systems. However, despite considering real-world deployment, the systematic investigation of challenges associated with applying RL in real-world industrial environments is not the primary focus of any of the mentioned studies.

Prior work has also investigated challenges associated with the real-world deployment of RL in other applications. \cite{dulac-arnoldChallengesRealworldReinforcement2021} identify and formalize a series of challenges in the real-world deployment of RL. The study does not consider a specific application area. \cite{schlegel2025towards} analyze the real-world challenges presented by \cite{dulac-arnoldChallengesRealworldReinforcement2021} for the application in power grids. \cite{schaferCrucialRoleProblem2025} demonstrate and discuss challenges associated with RL problem formulation choices and their effect on performance for the control of a helicopter testbed considering simulation-based and real-world training.

This work builds upon the real-world RL deployment presented by \cite{lademannRealWorldBenchmarkingControl2026}. We formulate the RL problem and identify the associated challenges. Finally, we analyze these challenges in the real-world application.

\end{document}